\documentclass[letterpaper, 10 pt, conference]{ieeeconf}  % Comment this line out if you need a4paper

\IEEEoverridecommandlockouts                              % This command is only needed if 
\usepackage[font={footnotesize}]{caption}
\usepackage{amsmath}
\usepackage[table,dvipsnames]{xcolor}
\usepackage{multirow}
\usepackage{booktabs}

\usepackage{graphicx}
\usepackage{hyperref}
\usepackage{adjustbox}
\usepackage{threeparttable}
\usepackage{amssymb}
\usepackage{subcaption}
\usepackage{blindtext}
\usepackage[absolute]{textpos}
\usepackage[normalem]{ulem}
\useunder{\uline}{\ul}{}

\usepackage{xcolor}
\hypersetup{
    colorlinks,
    linkcolor={red!70!black},
    citecolor={blue!50!black},
    urlcolor={blue!70!black}
}

\graphicspath{{./images/}}
\DeclareGraphicsExtensions{.pdf,.jpeg,.png}

\title{\LARGE \bf
SCOUT+: Towards Practical Task-Driven Drivers' Gaze Prediction
}

\author{Iuliia Kotseruba and John K. Tsotsos% <-this % stops a space
\thanks{The authors are with the Department of Electrical Engineering and Computer Science, York University, Toronto, Canada.
        Email: {\tt yulia84@yorku.ca, tsotsos@yorku.ca}}%
}

\begin{document}

\begin{textblock}{10}(1,0.5)
\noindent \footnotesize Accepted at IEEE Intelligent Vehicles Symposium (IV), 2024
\end{textblock}

\maketitle
\thispagestyle{empty}
\pagestyle{empty}

%%%%%%%%%%%%%%%%%%%%%%%%%%%%%%%%%%%%%%%%%%%%%%%%%%%%%%%%%%%%%%%%%%%%%%%%%%%%%%%%
\begin{abstract} 

Accurate prediction of drivers' gaze is an important component of vision-based driver monitoring and assistive systems. Of particular interest are safety-critical episodes, such as performing maneuvers or crossing intersections. In such scenarios, drivers' gaze distribution changes significantly and becomes difficult to predict, especially if the task and context information is represented implicitly, as is common in many state-of-the-art models. However, explicit modeling of top-down factors affecting drivers' attention often requires additional information and annotations that may not be readily available. 

In this paper, we address the challenge of effective modeling of task and context with common sources of data for use in practical systems. To this end, we introduce SCOUT+, a task- and context-aware model for drivers' gaze prediction, which leverages route and map information inferred from commonly available GPS data. We evaluate our model on two datasets, DR(eye)VE and BDD-A, and demonstrate that using maps improves results compared to bottom-up models and reaches performance comparable to the top-down model SCOUT which relies on privileged ground truth information. Code is available at \url{https://github.com/ykotseruba/SCOUT}.

\end{abstract}

%%%%%%%%%%%%%%%%%%%%%%%%%%%%%%%%%%%%%%%%%%%%%%%%%%%%%%%%%%%%%%%%%%%%%%%%%%%%%%%%
\section{Introduction}

Accurate modeling of drivers' gaze is necessary for vision-based monitoring and driver assistance systems. Historically, research in this area focused mainly on detecting drivers' distraction and other forms of inattention \cite{kotseruba2022practical}. Understanding whether attentive drivers adequately observe the scene also has a potential to improve road safety and may be used in driver training.  Of particular interest are episodes when drivers' actions may put them in the path of other road users, e.g., when navigating the intersections or performing lateral and longitudinal maneuvers. Intersections, for instance, comprise only a small fraction of the road network, but are disproportionately represented in the accident statistics \cite{2020_FHWA_Intersection_Safety, 2015_EU_Traffic_Safety}. According to \cite{shawky2020factors}, crashes at intersections are often linked to the drivers not observing the approaching road users due to neglect, occlusions, and distractions. The same study relates accidents during lane changes to recognition failures and lack of awareness, i.e., not looking at mirrors and blind spots, among other factors.

Although there is no complete model of visual attention during driving, a number of factors affecting how and when drivers look have been identified \cite{kotseruba2021behavioral}. In the psychological literature, various influences on attention are subdivided into bottom-up and top-down \cite{tsotsos2011computational}. Bottom-up or data-driven attention is involuntary attraction to unexpected or otherwise salient areas and events in the scene, e.g., flashing lights near railway crossings or pedestrians darting onto the road. Top-down or task-driven attention, as the name suggests, is guided by the demands of the current activity, for example, checking the rearview mirror before braking or looking both ways before making a turn at the intersection. 

Accurate models of driver attention should accommodate both types of influences, but few do so explicitly \cite{kotseruba2022practical}. In particular, it is difficult to take into account specific tasks or context in which they are performed by the driver because additional data is needed. For example, past attempts at including explicit top-down influences relied on precise vehicle information or numerous additional features and annotations \cite{borji2011computational,2021_T-ITS_Amadori,kotseruba2023understanding}.  As a result, partly due to lack of such annotations in the public datasets, many approaches instead learn correlations between the ground truth derived from human eye-tracking data and images of the scene \cite{palazzi2018predicting, 2018_ACCV_Xia, 2020_T-ITS_Deng, 2021_T-ITS_Fang, 2022_T-ITS_Gan}. 

In this paper, we address the issue of representing driver actions and surrounding context by using map and route information. We extend the SCOUT\footnote{SCOUT stands for Ta\textbf{S}k- and \textbf{C}ontext-m\textbf{O}d\textbf{U}lated a\textbf{T}tention.} model introduced in \cite{kotseruba2023understanding}, which relied on hand-coded text labels, e.g., for current/future tasks, types of intersections, etc. The proposed SCOUT+ instead infers relevant task and context information from the route and street network (in addition to images of the scene). To do so, we first infer street network and route from the available GPS data using the OpenStreetMap API. We then correct the route coordinates by matching them to the map, and rasterize both route and map for use in training and evaluation. The proposed model uses cross-attention to combine the encoded map and route representation with spatiotemporal visual features. Through experiments, we investigate the usefulness of the maps for representing the task and context elements. We evaluate SCOUT+ against SOTA models for video saliency and drivers' gaze prediction on two datasets, DR(eye)VE \cite{palazzi2018predicting} and BDD-A \cite{2018_ACCV_Xia}.

\section{Relevant Works}

Drivers' gaze prediction is a subfield of a larger research area dedicated to computational modeling of human visual attention. One of the long-standing problems is understanding the mechanisms that govern how people observe the scene: what areas they focus (fixate) on and why do they make overt eye movements (saccades) to look at something else. As mentioned earlier, in the literature, factors causing these changes in attention are subdivided into bottom-up (data-driven) and top-down (task-driven) \cite{tsotsos2011computational}. In driving, and other visually-guided activities, both types of influences are involved, but there is no comprehensive theory of the complex interactions between them \cite{sullivan2012role, hayhoe2014modeling, johnson2014predicting}.

\noindent
\textbf{Bottom-up models.} Many models for drivers' gaze prediction follow the bottom-up approach, i.e., learning a mapping from images or videos of the scene to human ground truth derived from eye-tracking data. Most such models share a similar encoder-decoder architecture, thus the main difference between them is in the types of inputs they use and how they process and combine different features. For example, BDD\nobreakdash-ANet \cite{2018_ACCV_Xia}, CDNN \cite{2020_T-ITS_Deng}, and ADA \cite{2022_T-ITS_Gan} use only image data, whereas other models take advantage of additional information, such as semantic maps \cite{palazzi2018predicting, 2020_CVPR_Pal, 2021_T-ITS_Fang}, optical flow \cite{palazzi2018predicting, hu2022novel}, and object detection \cite{2020_CVPR_Pal, 2022_T-ITS_Li}. Use of these features aims at improving scene understanding, for example, fine-grained motion analysis, distinguishing different types of road users, and establishing relationships between them and the ego-vehicle. Although undoubtedly useful, these features are only indirectly related to drivers' goals and actions.

\noindent
\textbf{Top-down models.} Fewer models incorporate explicit representations for drivers' actions. In early works, \cite{borji2011computational,borji2012probabilistic}, a variety of statistical modeling approaches (regression, nearest-neighbor clustering, SVMs, and Bayesian networks) were applied to predict likely gaze locations from a combination of image and action features. The latter were represented as a multidimensional vector encoding positions of the steering wheel and pedals, gear changes, and left/right signals. The authors showed that purely bottom-up models performed the worst, whereas inclusion of the task information, previous gaze location, and global context (gist of the scene \cite{oliva2005gist}) helped the most. In \cite{2017_IV_Tawari}, yaw rate (derived from GPS) was used to represent drivers' actions, such as turning, curve-following, and lane changes. Specifically, the authors proposed to compute gaze location priors as average saliency maps for binned yaw rate values. The priors were then multiplied with the final saliency map to modulate it, leading to a $5\%$ improvement on some metrics.

More recent approaches are HammerDrive \cite{2021_T-ITS_Amadori},  MEDIRL \cite{2021_ICCV_Baee}, and SCOUT \cite{kotseruba2023understanding}. HammerDrive consists of a 3D convolutional encoder for visual information and several LSTMs that output saliency maps for specific actions. Vehicle telemetry is used to determine current action and adjust contributions of the LSTMs to the final output. MEDIRL likewise tackles several specific tasks, such as merging-in, lane-keeping, and braking. The model relies on multiple encoders to capture task-related visual cues, such as objects, object proximity, road lanes, traffic lights, and brake lights. Inverse Reinforcement Learning (IRL) is then applied to learn the next best area to fixate on for the given action and context. SCOUT, in addition to drivers' tasks (maneuvers) also considers context, represented as text labels for the type of intersection and the ego-vehicle's right-of-way. The model has three components: a Transformer-based encoder for visual features, an encoder for text (speed, GPS coordinates, action labels) and context (proximity to intersection, intersection type, right-of-way, next action), and a multi-head attention fusion module and a decoder for modulating the visual features with task information.  

Although top-down models often show improvement over bottom-up ones, representing the task (even coarsely) requires additional information, which may be privileged or difficult to obtain. Towards making these systems more practical, here, we investigate usefulness of map and route information as a compact representation for surrounding context (upcoming intersections and road curves) and drivers' actions (speed and direction of travel). The main advantage of this approach is that information can be inferred from GPS, which is already available in some datasets and most consumer car models.

\section{Method}

\subsection{Problem formulation}

The problem of drivers' gaze prediction is formulated as follows: given a set of $n$ RGB images $I \in \mathbb{R}^{H\times W \times 3}$ recorded from the driver's perspective, predict where the driver is more likely to look in the last frame of the input set. Gaze is typically represented as a saliency map $S \in [0,1]^{H\times W}$, where pixels with higher values correspond to areas in the image which are more likely to attract attention.  

Following \cite{kotseruba2023understanding}, we consider task as a set of actions that lead to lateral and longitudinal changes in ego-vehicle motion, in addition to remaining stationary and maintaining lane and speed. Context includes the presence of an intersection, its type (e.g., signalized or unsignalized), and the ego-vehicle's priority with respect to other vehicles (right-of-way or yield).

\subsection{Map and route generation}
\label{sec:map_route}
\noindent
\textbf{Street network extraction.} Route is a set of GPS waypoints corresponding to the path driven by the ego-vehicle. Using the route, we extract map information as follows. We first obtain a street network for the geographical region using OpenStreetMap API \cite{OpenStreetMap} and OSMnx library \cite{boeing2017osmnx}. To reduce the size of the map, we plot only the driving network (omitting sidewalks and bike lanes) and crop the street network to the bounding box containing the route endpoints. To avoid loss of relevant street nodes, a small offset (500 to 1500 m) is added to the corners of the bounding box. 

\begin{figure}
\centering
\includegraphics[width=\columnwidth]{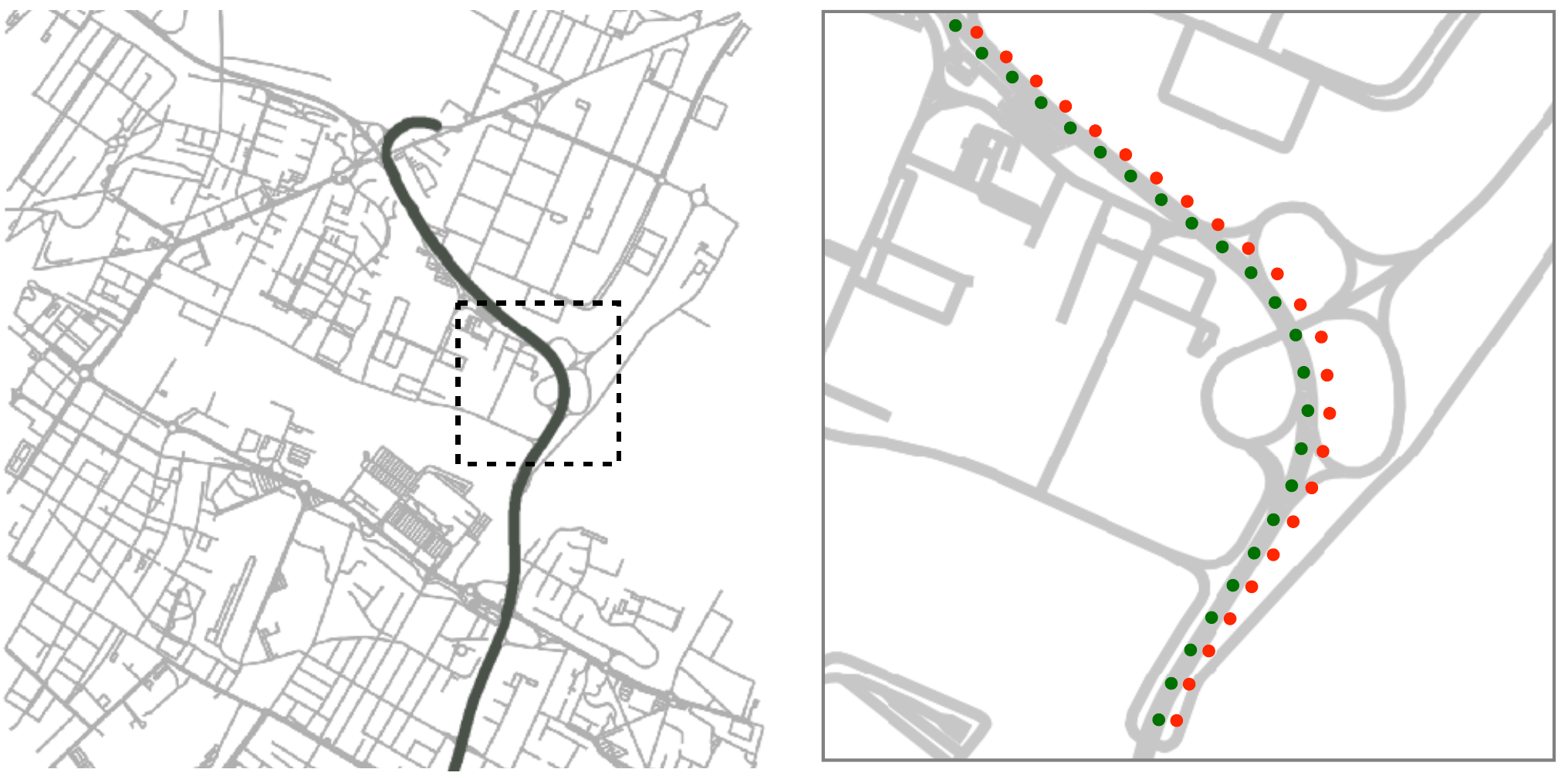}
\caption{Example of the map and route inferred from DR(eye)VE GPS data. Left: street network with overlaid route traveled by the vehicle. Right: enlarged portion of the map shows map matching the original noisy GPS coordinates (shown in red) to the street network (green). Map colors are inverted here for readability. For use in training and inference, street network and route are rendered as white lines on a black background.}
\label{fig:map_route}
\vspace{-1.5em}
\end{figure}

\begin{figure*}[htb]
\centering
\includegraphics[width=0.75\textwidth]{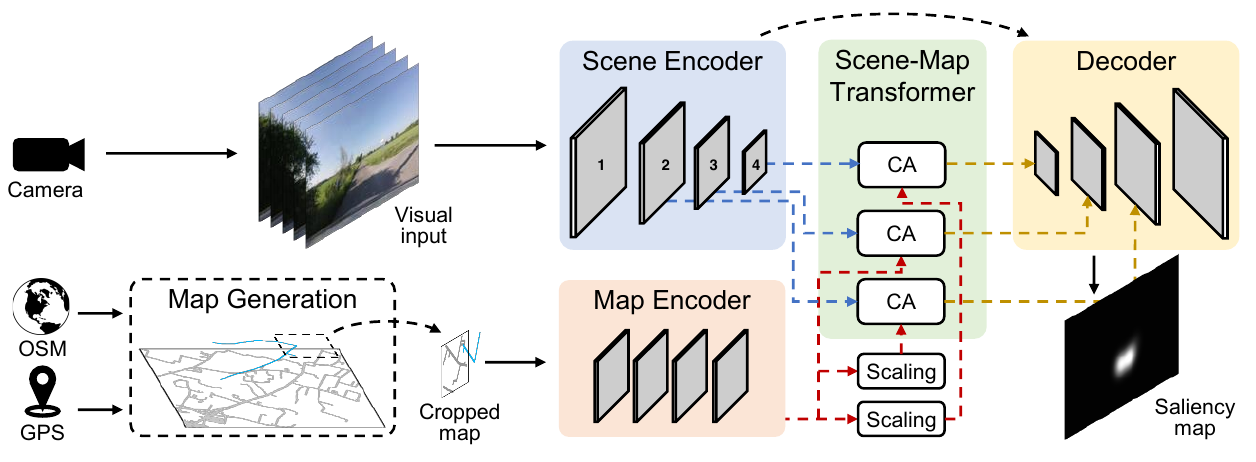}
\caption{Diagram of the SCOUT+ architecture.  \textit{Scene encoder} takes in a set of images and outputs 3D spatio-temporal features. To obtain map and route features, the \textit{map generation} module extracts the map and route information from the GPS data using OpenStreetMap (OSM) API. Then, a patch of the map corresponding to the location in the visual input is cropped and fed into the \textit{map encoder}, which produces 2D map and route features via a shallow CNN network. \textit{Scene-map transformer} applies cross-attention (CA) to fuse map and visual features. The \textit{decoder} receives input either directly from the encoder or from the CA blocks (shown as dashed arrows) and gradually mixes and upscales features to obtain the final saliency map.}
\label{fig:architecture}
\vspace{-1.5em}
\end{figure*}

\noindent
\textbf{GPS map matching.} GPS positional data is often inaccurate due to measurement and sampling errors \cite{chao2020survey}. As a result, route coordinates may deviate from the road boundaries. Map matching is a common technique that reduces positional errors by ``snapping'' GPS location coordinates onto the road network. Here, we use the implementation of the Hidden Markov map matching algorithm \cite{newson2009hidden} provided in the Valhalla\footnote{\url{https://github.com/valhalla/}} routing engine. Since GPS is sampled at 1 Hz, we interpolate the map matched coordinates to obtain a value for each frame in the video.

\noindent
\textbf{Map and route rasterization.} The final step is rasterizing the vector images of street network and route for use in training. We first plot the street graph as white lines on a black background. We then compute the size of the rasterized map image so that 1 px at 100 dpi resolution approximately corresponds to 1 m in the vector map. This is done to simplify cropping the map during training to the required lookahead distance in meters. Using this image size in pixels and latitude and longitude of the max and min coordinates in the vector map, we convert the GPS route coordinates to pixel locations in the image plane.

\noindent
\textbf{Sampling map and route.} During training and testing, we use a portion of the rasterized map and route. For each observation, we crop a square patch of the map centered at the last observed ego-vehicle location and with sides at a predefined lookahead distance away. We normalize orientations of all patches by rotating them such that the direction of travel for the given observation points up. This matches the map to the view of the scene and is similar to how the drivers might view maps on the navigation screen. To represent route, we create an empty patch of the same size and plot past observed locations of the vehicle corresponding to the input frames as a white line. Then we similarly rotate the patch with the plotted route to match the orientation of the patch with the map. For training and inference, route and map patches are concatenated along the depth dimension.

\subsection{Model architecture}

The proposed SCOUT+ model (shown in Figure \ref{fig:architecture}) consists of scene and map encoders, for visual spatio-temporal feature extraction and for map/route encoding, respectively, a scene-map cross-attention block, and a decoder that gradually fuses and upscales the features to produce the output. Below we discuss each module in detail.

\noindent
\textbf{Scene encoder.} As in \cite{kotseruba2023understanding} we use Video Swin Transformer (VST) \cite{2022_CVPR_Liu} for spatio-temporal feature extraction from a stack of RGB input frames. Given a video clip input $\mathcal{I}=\{{I_t}\}^{T}_{t=1}$, the VST visual backbone produces $N$ visual feature maps $\mathcal{F}^{v}=\{{f^{v}_n}\}^{N}_{n=1}$. Here, we use the modified VST proposed in \cite{botach2022end}, which does not downsample the input temporally. The outputs of all four encoder blocks are used.

\noindent
\textbf{Map encoder.} To efficiently encode map and route information, we design a lightweight CNN, following similar approaches in the self-driving domain \cite{amirloo2022latentformer,salzmann2020trajectron}. The network is composed of four 2D convolutional layers with 10, 20, 10, and 1 kernels of sizes $5\times 5$, $3\times 3$, $3\times 3$, and $1\times 1$, respectively, each followed by a leaky ReLU activation \cite{maas2013rectifier}. The network has a fixed input of size $M \times H_m \times W_m$, where the number of channels $M$ corresponds to the number of map features. This 2D encoding is then rescaled via bilinear interpolation and replicated along other dimensions to match the sizes of the scene encoder outputs, resulting in $N$ map features $\mathcal{F}^{m}=\{{f^{m}_n}\}^{N}_{n=1}$.

%Task and context input consists of per-sample text labels and per-frame numeric values (indicated as $l_1, \dots l_n$ and $n_1, \dots, n_m$ in Figure \ref{fig:architecture}), representing \textit{global context} (weather, time of day, and location labels provided with DR(eye)VE), \textit{local context} (next lateral action label, distance to next intersection (m) derived from GPS, and priority label), and \textit{current action} (speed in km/h, acceleration in m/s\textsuperscript{2}, and lateral action label). We use a linear embedding layer to encode text labels and downsample the numeric values along the temporal dimension to match the temporal dimension of VST outputs. We then stack the encoded features, apply an affine transformation to match the feature dimension $C$ of the corresponding encoder block and replicate the features across that block's spatial dimensions. 

\noindent
\textbf{Scene-map transformer.} We use cross-attention (CA) \cite{2017_NeurIPS_Vaswani} to fuse visual features with map and route information as follows. The encoded visual features ${f^{v}_n}$ are used as key and value, and the  map encoder output ${f^{m}_n}$ as query in
%\begin{equation*}
$CA(f^{v}_n, f^{m}_n)=Softmax(\frac{f^{m}_n W^Q \cdot (f^{v}_n W^K)^T}{\sqrt{d_{head}}})f^{v}_n W^V$,
%\end{equation*}
%\noindent
where $W^Q$, $W^K$, and $W^V$ $\in \mathcal{R}^{c\times d_{head}}$ are learnable parameters.

\noindent
\textbf{Decoder.} The decoder structure follows \cite{kotseruba2023understanding} and consists of four blocks of upsampling, ReLU and 3D convolutional layers. Decoder blocks receive outputs of the encoder or modulation blocks, increase their spatial dimension via upsampling, pass them through ReLU activation layers, and reduce their depth using a 3D convolution layer. The final output is passed through the sigmoid activation function.

\begin{table*}[!htp]
\centering
\caption{Evaluation results on the DR(eye)VE dataset. $\uparrow$ and $\downarrow$ indicate that larger and smaller values are better. Best and second-best values are shown as \textbf{bold} and \underline{underlined}, respectively. \textit{Task}---whether task/context features are used, \textit{Map}---whether map (without route) is used, \textit{Enc. block}---at which encoder block map/task features are fused with visual information. The following abbreviations are used for actions and context subsets: \textit{None}---maintain speed/lane, \textit{Stop}---ego-vehicle is stopped, \textit{Acc}---longitudinal acceleration, \textit{Dec}---longitudinal deceleration, \textit{Lat}---lateral actions only, \textit{Lat/Lon}---simultaneous lateral and longitudinal actions, \textit{RoW}---ego-vehicle has right-of-way.}
\vspace{-1em}
\setlength{\tabcolsep}{0.25em}

\resizebox{\textwidth}{!}{%
\begin{tabular}{ccccccccccccccccccccccccc}
 &
   &
   &
   &
   &
   &
   &
   &
   &
   &
   &
   &
   &
   &
   &
   &
   &
  \multicolumn{8}{c}{Context (KLD↓)} \\ \cline{18-25} 
 &
   &
   &
   &
   &
  \multicolumn{4}{c}{All} &
   &
  \multicolumn{6}{c}{Actions (KLD↓)} &
   &
  \multicolumn{2}{c}{Signalized} &
  \multicolumn{2}{c}{Unsignalized} &
  \multicolumn{2}{c}{Roundabout} &
  \multicolumn{2}{c}{Highway} \\ \cline{6-9} \cline{11-16} \cline{18-25} 
 &
   &
  Task &
  Map &
  Enc. block &
  KLD↓ &
  CC↑ &
  NSS↑ &
  SIM↑ &
   &
  None &
  Acc &
  Dec &
  Lat &
  Lat/Lon &
  Stop &
   &
  RoW &
  Yield &
  RoW &
  Yield &
  RoW &
  Yield &
  RoW &
  Yield \\ \cline{1-9} \cline{11-16} \cline{18-25} 
\multicolumn{2}{c|}{DReyeVENet} &
  - &
  - &
  \multicolumn{1}{c|}{-} &
  2.13 &
  \textbf{0.67} &
  2.01 &
  \textbf{0.56} &
   &
  1.58 &
  1.83 &
  2.68 &
  4.18 &
  4.27 &
  3.38 &
   &
  2.42 &
  4.49 &
  2.30 &
  7.72 &
  3.16 &
  7.50 &
  1.39 &
  8.10 \\
\multicolumn{2}{c|}{CDNN} &
  - &
  - &
  \multicolumn{1}{c|}{-} &
  2.00 &
  \textbf{0.67} &
  2.03 &
  0.54 &
   &
  1.39 &
  1.53 &
  2.09 &
  3.40 &
  3.88 &
  5.78 &
   &
  2.43 &
  3.23 &
  1.77 &
  6.19 &
  2.66 &
  4.69 &
  1.24 &
  7.37 \\
\multicolumn{2}{c|}{BDD-ANet} &
  - &
  - &
  \multicolumn{1}{c|}{-} &
  1.61 &
  0.64 &
  2.05 &
  0.48 &
   &
  1.22 &
  1.44 &
  1.77 &
  2.70 &
  3.09 &
  3.19 &
   &
  1.91 &
  2.42 &
  1.69 &
  4.28 &
  2.56 &
  3.92 &
  1.05 &
  6.01 \\
\multicolumn{2}{c|}{ViNet} &
  - &
  - &
  \multicolumn{1}{c|}{-} &
  1.21 &
  {\ul 0.66} &
  1.99 &
  0.48 &
   &
  0.96 &
  1.11 &
  1.33 &
  1.76 &
  2.08 &
  2.43 &
   &
  1.46 &
  2.11 &
  1.29 &
  2.59 &
  1.78 &
  2.67 &
  0.89 &
  {\ul 1.99} \\ \cline{1-9} \cline{11-16} \cline{18-25} 
\multicolumn{2}{c|}{SCOUT w/o task} &
  - &
  - &
  \multicolumn{1}{c|}{} &
  1.07 &
  0.63 &
  3.89 &
  0.52 &
   &
  0.81 &
  0.99 &
  1.25 &
  1.61 &
  1.99 &
  2.43 &
   &
  1.15 &
  2.46 &
  1.10 &
  3.11 &
  \textbf{1.26} &
  2.72 &
  0.74 &
  2.24 \\
\multicolumn{2}{c|}{SCOUT w/task} &
  + &
  - &
  \multicolumn{1}{c|}{2} &
  \textbf{0.92} &
  \textbf{0.67} &
  \textbf{4.17} &
  {\ul 0.55} &
   &
  {\ul 0.70} &
  {\ul 0.86} &
  \textbf{1.05} &
  \textbf{1.40} &
  \textbf{1.62} &
  2.03 &
   &
  \textbf{0.99} &
  {\ul 1.60} &
  \textbf{0.89} &
  \textbf{2.09} &
  {\ul 1.43} &
  \textbf{2.03} &
  {\ul 0.67} &
  \textbf{1.95} \\ \cline{1-9} \cline{11-16} \cline{18-25} 
\multicolumn{2}{c|}{\multirow{4}{*}{SCOUT+ (ours)}} &
  - &
  + &
  \multicolumn{1}{c|}{2,3,4} &
  0.96 &
  {\ul 0.66} &
  4.09 &
  0.53 &
   &
  0.73 &
  0.87 &
  1.11 &
  1.54 &
  1.95 &
  \textbf{1.93} &
   &
  {\ul 1.07} &
  1.79 &
  1.00 &
  2.82 &
  1.71 &
  2.49 &
  \textbf{0.65} &
  2.22 \\
\multicolumn{2}{c|}{} &
  - &
  + &
  \multicolumn{1}{c|}{2} &
  0.99 &
  {\ul 0.66} &
  4.06 &
  0.54 &
   &
  0.74 &
  0.92 &
  1.18 &
  1.55 &
  1.92 &
  2.04 &
   &
  1.13 &
  1.85 &
  1.03 &
  2.81 &
  1.60 &
  2.82 &
  0.69 &
  2.25 \\
\multicolumn{2}{c|}{} &
  - &
  + &
  \multicolumn{1}{c|}{3} &
  {\ul 0.94} &
  \textbf{0.67} &
  {\ul 4.15} &
  {\ul 0.55} &
   &
  \textbf{0.69} &
  \textbf{0.85} &
  {\ul 1.08} &
  1.55 &
  1.86 &
  {\ul 2.00} &
   &
  1.09 &
  \textbf{1.57} &
  0.97 &
  2.56 &
  1.53 &
  2.43 &
  \textbf{0.65} &
  2.16 \\
\multicolumn{2}{c|}{} &
  - &
  + &
  \multicolumn{1}{c|}{4} &
  0.95 &
  \textbf{0.67} &
  4.12 &
  0.53 &
   &
  {\ul 0.70} &
  {\ul 0.86} &
  {\ul 1.08} &
  {\ul 1.46} &
  {\ul 1.73} &
  2.22 &
   &
  {\ul 1.07} &
  1.66 &
  {\ul 0.93} &
  {\ul 2.45} &
  1.53 &
  {\ul 2.38} &
  {\ul 0.67} &
  2.10 \\ \cline{1-9} \cline{11-16} \cline{18-25} 
\end{tabular}%
}

\label{tab:DReyeVE_results}
\vspace{-1em}
\end{table*}

\section{Evaluation setup}
\subsection{Data, models, and metrics}

We evaluate our model on two popular datasets for drivers' gaze prediction: DR(eye)VE \cite{palazzi2018predicting} and BDD-A \cite{2018_ACCV_Xia}. For training and testing models on DR(eye)VE, we use the corrected and annotated ground truth saliency maps from \cite{kotseruba2023understanding}. We also annotate BDD-A videos with the same task and context labels for this work.

Performance comparisons are made to a representative set of the models for which official training code is available. Specifically, we use the driver gaze prediction models that take as input a single image (CDNN \cite{2016_T-ITS_Deng}) or a stack of images (DReyeVENet \cite{palazzi2018predicting}, BDD-ANet \cite{2018_ACCV_Xia}, and SCOUT \cite{kotseruba2023understanding}). In addition, we use a bottom-up generic video saliency model, ViNet \cite{2021_IROS_Jain}. The models are trained on each dataset with their default hyperparameters.

The results are reported with respect to the following common saliency metrics: Kullback-Leibler divergence (KLD), Pearson's correlation coefficient (CC), normalized scanpath saliency (NSS), and histogram similarity (SIM) \cite{2018_PAMI_Bylinskii}. The Python implementations of all metrics are provided in \cite{fahimi2021metrics}. 

\subsection{Implementation}
The input to the proposed model is a set of RGB video frames and a single map/route patch. The size of the RGB input is fixed to 16 consecutive frames ($\approx 0.5$s observation) each resized to $224\times224\times3$ px. We use Swin-S VST model pre-trained on the Kinetics-400 dataset \cite{2017_CVPR_Carreira}. As described earlier in Section \ref{sec:map_route}, map/route input is cropped based on the map radius setting (between 25 and 200 m), rotated, and resized to a fixed size of $128\times128$ px. Route features are appended to the map patch along the channel dimension. The weights of the pre-trained VST backbone are fine-tuned during training. The map-scene cross-attention modules have 2 attention heads and embedding sizes set to the channel dimensions of the corresponding VST outputs. 

We use the first 34 videos from the DR(eye)VE dataset for training, videos 35--37 for validation, and the remaining videos for testing. From BDD-A, we exclude 135 videos, those with quality issues (e.g., unfocused or tilted camera) and ones without gaze or GPS data. Training samples are generated by splitting the videos into 16-frame segments with 8 frame overlap. U-turns and reversing are very rare in all datasets and gaze data is inaccurate in most. Thus, samples containing these events were excluded from training, validation, and testing. The model is trained on a single NVIDIA Titan 1080Ti GPU for 20 epochs with the KLD loss, Adam optimizer \cite{2014_arXiv_2014}, early stopping based on validation loss, a constant learning rate of $1e-4$, and batch size of $4$.

\section{Evaluation results}
\subsection{Results on the DR(eye)VE dataset}
We first established the optimal map size by running tests with different amount of lookahead, from 50 and up to 200 meters, to represent different planning horizons. We found that adding maps of any size was beneficial compared to using only the visual information. Gradually expanding the map around the ego-vehicle improved the results, but saturated at 150 m, and started to decline at 200 m. The best improvements were achieved at 100 m, therefore the map size was fixed at this value for the rest of the experiments.

\begin{table}[!tp]
\centering
\caption{Effect of different inputs on SCOUT+ performance on DR(eye)VE. All models are trained with 100 m context. Models that use visual and map information, fuse it with the last encoder block.}
\vspace{-0.5em}
\setlength{\tabcolsep}{0.45em}

\resizebox{0.8\columnwidth}{!}{%
\begin{tabular}{ccc|cccc}
Scene & Map & Route & KLD↓          & CC↑           & NSS↑          & SIM↑          \\ \hline
      & +   &       & 1.14          & 0.62          & 3.84          & 0.47          \\
+     & +   &       & \textbf{0.95} & \textbf{0.67} & \textbf{4.12} & 0.53          \\
+     & +   & +     & 0.97          & 0.66          & 4.10          & \textbf{0.54} \\ \hline
\end{tabular}%
}

\label{tab:DReyeVE_map_features}
\vspace{-0.5em}
\end{table}

\noindent
\textbf{Overall performance.} Table \ref{tab:DReyeVE_results} shows the results of SCOUT+ on DR(eye)VE compared to performance of the state-of-the-art models and two versions of SCOUT---with and without task information. Overall, we observe that all versions of SCOUT+ with map information result in improvement over the SCOUT model without task and are comparable to SCOUT that has access to ground truth task and context information. For example, SCOUT+ performs as well as SCOUT with task on SIM and CC, and is comparable on NSS and KLD, which are more sensitive to false positives.

\begin{table*}[!htp]
\centering
\caption{Evaluation results on the BDD-A dataset. $\uparrow$ and $\downarrow$ indicate that larger and smaller values are better. Best and second-best values are shown as \textbf{bold} and \underline{underlined}, respectively. Abbreviations for actions and context column titles are the same as in Table \ref{tab:DReyeVE_results}.}
\vspace{-0.5em}

\setlength{\tabcolsep}{0.25em}

\resizebox{0.85\textwidth}{!}{%
\begin{tabular}{ccccccccccccccccccccc}
 &
   &
   &
   &
   &
   &
   &
   &
   &
   &
   &
   &
   &
   &
   &
   &
   &
  \multicolumn{4}{c}{Context (KLD↓)} \\ \cline{18-21} 
 &
   &
   &
   &
   &
  \multicolumn{4}{c}{All} &
   &
  \multicolumn{6}{c}{Actions (KLD↓)} &
   &
  \multicolumn{2}{c}{Signalized} &
  \multicolumn{2}{c}{Unsignalized} \\ \cline{6-9} \cline{11-16} \cline{18-21} 
 &
   &
  Task &
  Map &
  Enc. block &
  KLD↓ &
  CC↑ &
  NSS↑ &
  SIM↑ &
   &
  None &
  Acc &
  Dec &
  Lat &
  Lat/Lon &
  Stop &
   &
  RoW &
  Yield &
  RoW &
  Yield \\ \cline{1-9} \cline{11-16} \cline{18-21} 
\multicolumn{2}{c|}{DReyeVENet} &
  - &
  - &
  \multicolumn{1}{c|}{-} &
  1.84 &
  0.62 &
  1.61 &
  0.46 &
   &
  1.81 &
  1.79 &
  1.71 &
  2.91 &
  2.25 &
  2.13 &
   &
  1.88 &
  3.19 &
  1.78 &
  3.07 \\
\multicolumn{2}{c|}{CDNN} &
  - &
  - &
  \multicolumn{1}{c|}{-} &
  1.84 &
  0.51 &
  1.30 &
  0.39 &
   &
  1.85 &
  1.78 &
  1.79 &
  1.97 &
  1.85 &
  2.16 &
   &
  1.81 &
  1.96 &
  1.60 &
  2.01 \\
\multicolumn{2}{c|}{BDD-ANet} &
  - &
  - &
  \multicolumn{1}{c|}{-} &
  1.43 &
  {\ul 0.64} &
  1.53 &
  \textbf{0.50} &
   &
  1.42 &
  1.39 &
  1.39 &
  1.61 &
  1.54 &
  1.60 &
   &
  1.57 &
  2.26 &
  1.43 &
  1.83 \\
\multicolumn{2}{c|}{ViNet} &
  - &
  - &
  \multicolumn{1}{c|}{-} &
  1.04 &
  \textbf{0.66} &
  1.58 &
  0.47 &
   &
  0.97 &
  1.00 &
  1.03 &
  1.17 &
  1.15 &
  \textbf{1.25} &
   &
  \textbf{1.15} &
  {\ul 1.30} &
  1.03 &
  1.27 \\ \cline{1-9} \cline{11-16} \cline{18-21} 
\multicolumn{2}{c|}{SCOUT w/o task} &
  - &
  - &
  \multicolumn{1}{c|}{-} &
  1.04 &
  0.63 &
  3.06 &
  0.48 &
   &
  0.99 &
  0.98 &
  1.03 &
  {\ul 1.11} &
  1.07 &
  1.29 &
   &
  1.18 &
  1.49 &
  1.01 &
  1.25 \\
\multicolumn{2}{c|}{SCOUT w/task} &
  + &
  - &
  \multicolumn{1}{c|}{2} &
  {\ul 1.02} &
  0.63 &
  {\ul 3.08} &
  {\ul 0.49} &
   &
  0.99 &
  \textbf{0.95} &
  {\ul 1.02} &
  \textbf{1.05} &
  \textbf{1.05} &
  1.30 &
   &
  1.18 &
  \textbf{1.26} &
  {\ul 0.97} &
  \textbf{1.20} \\ \cline{1-9} \cline{11-16} \cline{18-21} 
\multicolumn{2}{c|}{\multirow{4}{*}{SCOUT+ (ours)}} &
  - &
  + &
  \multicolumn{1}{c|}{2,3,4} &
  1.10 &
  0.60 &
  2.90 &
  0.45 &
   &
  1.06 &
  1.05 &
  1.08 &
  1.29 &
  1.19 &
  1.34 &
   &
  1.24 &
  1.53 &
  1.07 &
  1.37 \\
\multicolumn{2}{c|}{} &
  - &
  + &
  \multicolumn{1}{c|}{2} &
  1.05 &
  0.62 &
  3.01 &
  0.47 &
   &
  1.02 &
  1.00 &
  1.04 &
  1.08 &
  1.08 &
  1.31 &
   &
  1.28 &
  \textbf{1.26} &
  1.05 &
  1.25 \\
\multicolumn{2}{c|}{} &
  - &
  + &
  \multicolumn{1}{c|}{3} &
  \textbf{1.01} &
  0.63 &
  {\ul 3.08} &
  {\ul 0.49} &
   &
  {\ul 0.98} &
  \textbf{0.95} &
  \textbf{1.00} &
  \textbf{1.05} &
  {\ul 1.06} &
  1.27 &
   &
  {\ul 1.17} &
  1.33 &
  \textbf{0.94} &
  \textbf{1.20} \\
\multicolumn{2}{c|}{} &
  - &
  + &
  \multicolumn{1}{c|}{4} &
  \textbf{1.01} &
  {\ul 0.64} &
  \textbf{3.10} &
  0.48 &
   &
  \textbf{0.96} &
  {\ul 0.96} &
  \textbf{1.00} &
  {\ul 1.11} &
  1.14 &
  {\ul 1.26} &
   &
  1.19 &
  1.34 &
  1.01 &
  {\ul 1.24} \\ \cline{1-9} \cline{11-16} \cline{18-21} 
\end{tabular}%
}
\label{tab:BDDA_results}
\vspace{-1em}
\end{table*}

\begin{table}[!tp]
\centering
\caption{Effect of different inputs on SCOUT+ performance on BDD-A. All models are trained with 100 m context. Models that use visual and map information, fuse it with the last encoder block.}
\vspace{-0.5em}
\setlength{\tabcolsep}{0.45em}

\resizebox{0.8\columnwidth}{!}{%
\begin{tabular}{ccc|llll}
Scene & Map & Route & \multicolumn{1}{c}{KLD↓} & \multicolumn{1}{c}{CC↑} & \multicolumn{1}{c}{NSS↑} & \multicolumn{1}{c}{SIM↑} \\ \hline
  & + &   & 1.57          & 0.43          & 2.08          & 0.33          \\
+ & + &   & \textbf{1.01} & \textbf{0.63} & \textbf{3.08} & \textbf{0.49} \\
+ & + & + & 1.03          & \textbf{0.63} & 3.07          & \textbf{0.49} \\ \hline
\end{tabular}%
}

\label{tab:BDDA_map_features}
\vspace{-1.5em}
\end{table}

\begin{figure}[!htb]
\centering
\includegraphics[width=\columnwidth]{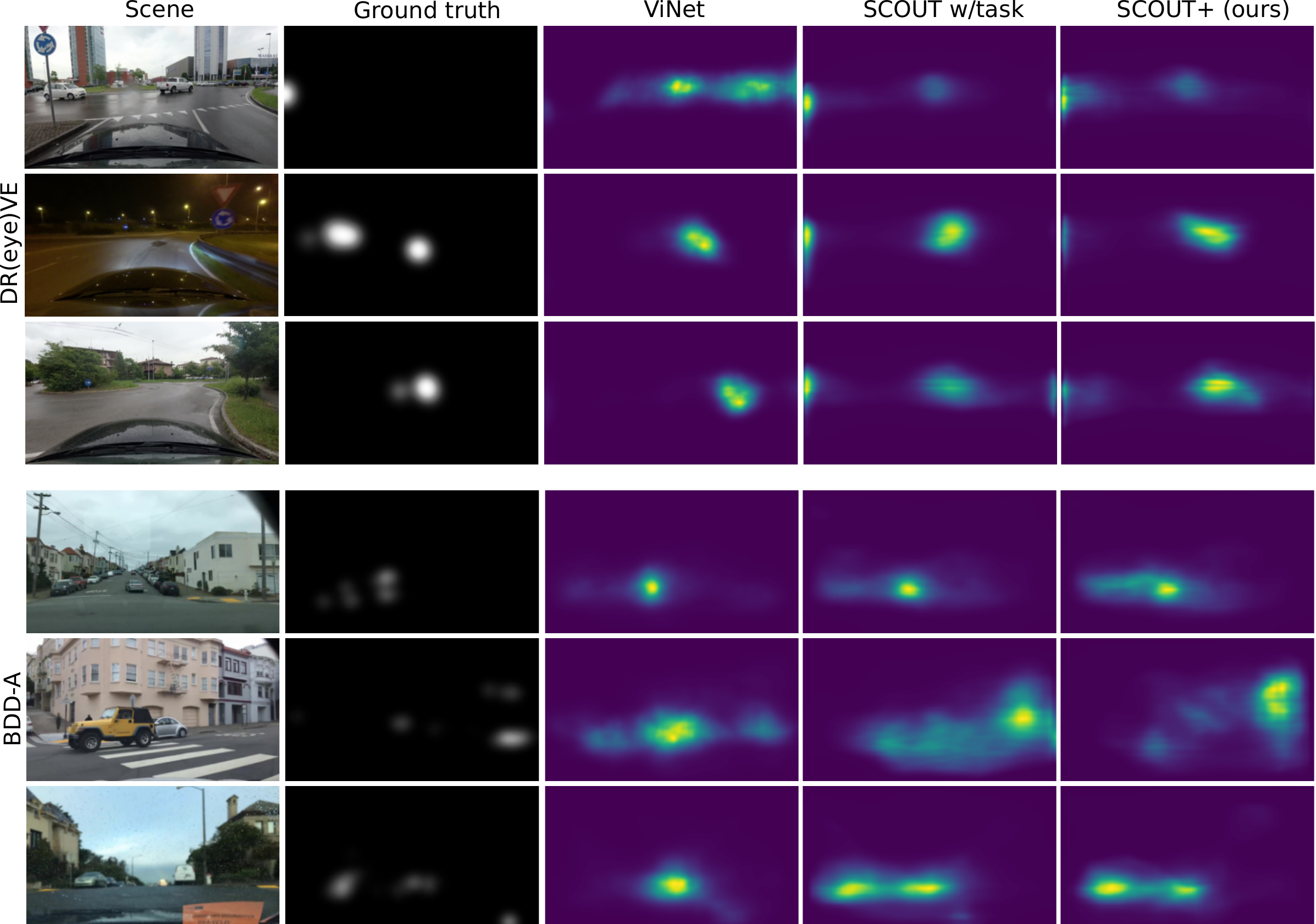}
\caption{Qualitative samples showing performance of SCOUT+ on challenging scenarios near intersections in DR(eye)VE (top 3 rows) and BDD-A (bottom 3 rows) against the bottom-up model ViNet and SCOUT which uses privileged task information.}
\label{fig:qual_samples}
\vspace{-2em}
\end{figure}

In addition, several versions of SCOUT+ are shown where map is combined with outputs from different blocks of the scene encoder. Although differences in performance on the entire dataset are relatively small (up to 2\%) across versions, it is evident that adding map to features from later encoder blocks is better and that fusing features from all blocks is not as effective. We speculate that spatio-temporal information in the early encoder blocks may be too low-level, thus its exclusion reduces noise. At the same time, regardless of how fusion is done, maps lead to significant improvements on challenging subsets corresponding to lateral actions and intersections. For example, decelerations and lateral maneuvers often occur near intersections, for which map information is the most useful. In particular, yielding scenarios and lateral actions see the largest gains (up to 30\% KLD) compared to the no-task version of SCOUT. SCOUT+ accurately predicts gaze of drivers who enter roundabouts or intersections (Figure \ref{fig:qual_samples} rows 1,2) and check for conflicting traffic, sometimes, even if the driver did not check (row 3). Not all action and context categories see equal improvements, but it is difficult to associate them with specific map or route features due to diversity of the scenarios in each subset.

\noindent
\textbf{Effect of different map features.} To further investigate the effect of maps, we train the model on only map information, i.e., with no visual input. As can be seen in Table \ref{tab:DReyeVE_map_features}, despite a predictable performance decrease across all metrics, the results are still competitive with some bottom-up models. Besides confirming the usefulness of maps, it is likely that the properties of the DR(eye)VE dataset contribute to this outcome. Since nearly half of the videos in DR(eye)VE are collected in rural or nighttime scenes with little traffic, the properties of the road, which are sufficiently represented by the street network, become more dominant. On the other hand, addition of the observed trajectory (route), which provides representation for the ego-vehicle's direction and speed of motion, leads to only minor improvements. We speculate, that to achieve more significant performance gains, a more precise and complete map representation is needed. For example, for maneuvers, one needs to know which lane is occupied by the ego-vehicle and whether other road users are present.

\subsection{Results on the BDD-A dataset}
We report evaluation results on BDD-A in Table \ref{tab:BDDA_results}. Note that here even using ground truth task information (SCOUT w/task) does not lead to improvements of the same magnitude as in the case of DR(eye)VE. There are several reasons for this. First, BDD-A is composed of short 10 s clips cropped around braking events from the BDDV dataset \cite{xu2017end}, where drivers' actions are reactive rather than planned. Second, many clips terminate before or in the middle of the planned actions, e.g., lane changes and crossing the intersections. Third, unlike DR(eye)VE, eye-tracking data for BDD\nobreakdash-A  was collected in the lab with observers viewing driving footage on the monitor. Finally, BDD-A ground truth combines gaze of at least four subjects, whereas in DR(eye)VE there is only one subject per video. A combination of these factors leads to qualitatively different gaze distributions in BDD-A. For example, gaps in performance between \textit{maintain} and \textit{lateral} actions and different types of intersections are not as large for this dataset compared to DR(eye)VE, which warrants further investigation.

%On the action subsets, there is consistent improvement on deceleration actions. In BDD-A, a large portion of the deceleration actions is stopping before intersections, so the map information is likely of help here. However, the majority of lateral actions are lane changes midblock, for which map fidelity is insufficient.

We repeat the same set of experiments as before to evaluate the proposed SCOUT+ model. Here also, map information leads to performance improvement, particularly, when it is fused with later encoder blocks (Table \ref{tab:BDDA_results}). As before, relying only on maps without any visual input hinders performance (Table \ref{tab:BDDA_map_features}). However, the decrease is more significant in relative terms, likely because in BDD-A there are more interactions with other road users, thus road features alone are not as helpful. Nevertheless, as shown in Figure \ref{fig:qual_samples}, SCOUT+ learns to anticipate drivers' actions, such as left (rows 4,6) and right (row 5) turns at unsignalized intersections, similar to SCOUT that uses ground truth task information. Note that in these scenarios, the bottom-up model ViNet remains more centered. 

\section{Conclusions}

In this paper, we investigated modeling task and context influences on drivers' attention for practical systems. We proposed SCOUT+ that utilized map information in an encoder-decoder architecture with cross-attention fusion blocks for combining visual and map inputs. While other top-down models rely on extensive annotations or privileged information, we show that a coarse map information derived from GPS data is a viable alternative since it is available in public datasets and most consumer vehicle models. Extensive experimentation on two datasets against several state-of-the-art models shows that our method achieves competitive results, even compared to the previous top-performing method that used ground truth data. In particular, with map, significant performance improvements are achieved on the challenging and potentially safety-critical scenarios involving lateral actions and intersections. The main limitation of the maps in the current implementation is their low fidelity. We believe that adding locations of lanes and other road users, types of intersections, and more accurate and detailed vehicle sensor data would help further improve the results.


%%%%%%%%%%%%%%%%%%%%%%%%%%%%%%%%%%%%%%%%%%%%%%%%%%%%%%%%%%%%%%%%%%%%%%%%%%%%%%%%

%%%%%%%%%%%%%%%%%%%%%%%%%%%%%%%%%%%%%%%%%%%%%%%%%%%%%%%%%%%%%%%%%%%%%%%%%%%%%%%%

%%%%%%%%%%%%%%%%%%%%%%%%%%%%%%%%%%%%%%%%%%%%%%%%%%%%%%%%%%%%%%%%%%%%%%%%%%%%%%%%

\noindent
\textbf{Acknowledgement.} Work supported by the Air Force Office of Scientific Research under award number FA9550-22-1-0538 (Computational Cognition and Machine Intelligence, and Cognitive and Computational Neuroscience portfolios); the Canada Research Chairs Program (950-231659); Natural Sciences and Engineering Research Council of Canada (RGPIN-2022-04606).

%%%%%%%%%%%%%%%%%%%%%%%%%%%%%%%%%%%%%%%%%%%%%%%%%%%%%%%%%%%%%%%%%%%%%%%%%%%%%%%%
\bibliographystyle{IEEEtran}
\bibliography{references}

\end{document}